\documentclass[10pt,conference,a4paper]{IEEEtran}
\pdfoutput=1

\usepackage{cite}
\usepackage{amsmath,amssymb,amsfonts}
\usepackage{algorithmic}
\usepackage{multirow}
\usepackage[pdftex]{graphicx}

\begin{document}
%
\title{A Prototype-Based Generalized Zero-Shot Learning Framework for Hand Gesture Recognition}

\author{\IEEEauthorblockN{Jinting Wu\IEEEauthorrefmark{1}\IEEEauthorrefmark{2}, Yujia Zhang\IEEEauthorrefmark{1}, Xiaoguang Zhao\IEEEauthorrefmark{1}}
\IEEEauthorblockA{\IEEEauthorrefmark{1}\textit{The State Key Laboratory of Management and Control for Complex Systems} \\
\textit{Institute of Automation, Chinese Academy of Sciences}\\
\IEEEauthorrefmark{2}\textit{School of Artificial Intelligence}\\
\textit{University of Chinese Academy of Sciences}\\
Beijing, China \\
Email: \{wujinting2016, zhangyujia2014, xiaoguang.zhao\}@ia.ac.cn}
}


\maketitle

\begin{abstract}
Hand gesture recognition plays a significant role in human-computer interaction for understanding various human gestures and their intent. However, most prior works can only recognize gestures of limited labeled classes and fail to adapt to new categories. The task of Generalized Zero-Shot Learning (GZSL) for hand gesture recognition aims to address the above issue by leveraging semantic representations and detecting both seen and unseen class samples. In this paper, we propose an end-to-end prototype-based GZSL framework for hand gesture recognition which consists of two branches. The first branch is a prototype-based detector that learns gesture representations and determines whether an input sample belongs to a seen or unseen category. The second branch is a zero-shot label predictor which takes the features of unseen classes as input and outputs predictions through a learned mapping mechanism between the feature and the semantic space. We further establish a hand gesture dataset that specifically targets this GZSL task, and comprehensive experiments on this dataset demonstrate the effectiveness of our proposed approach on recognizing both seen and unseen gestures.
\end{abstract}


%
\IEEEpeerreviewmaketitle

\section{Introduction}
Hand gesture recognition has been widely applied in various fields, such as post-stroke rehabilitation\cite{rehabilitation}, sign language recognition\cite{signlanguage}, emotion recognition\cite{emotion} and human-robot interaction\cite{HRI}. However, most existing works can only recognize a limited number of categories that have been seen during training and fail to extend to new categories. Besides, the high accuracy of hand gesture recognition requires a large number of labeled data from different classes, which in practice is costly and time-consuming. Therefore, it is critical to transfer the learned knowledge from seen to unseen categories and recognize unseen hand gesture classes, in order to better understand the intent of a user's new hand gesture.

Zero-Shot Learning (ZSL), where the goal is to accurately recognize data of unseen classes, provides a solution for tackling the above challenges. ZSL methods establish associations between seen and unseen categories with side information such as attributes\cite{firstZSL,SAE} and semantic vectors\cite{CONSE}. Note that in ZSL tasks, training and test classes are strictly disjoint, and only data from the seen categories are used during training. This may lead to inferior performance due to the inherent bias towards the seen classes when data from both seen and unseen classes are available at test time. In other words, the classifier tends to misidentify the samples from unseen categories as seen categories\cite{action}. To solve this problem, GZSL, a more general task is proposed, where samples from the unseen categories are mixed with the seen categories in the test set\cite{review}. It aims to reduce the effect of such bias towards seen classes in a less restricted setting where training and testing categories are not disjoint.

Although the ZSL and GZSL approaches for object recognition\cite{firstZSL,SAE,ESZSL,CADA-VAE,f-CLSWGAN,SCoRe} have been largely investigated and achieved great success, approaches that target the ZSL/GZSL for dynamic hand gesture recognition are less explored. Thomason and Knepper\cite{HRI} first developed a ZSL gesture recognition system to understand user's intent by leveraging coordinated natural language, gesture, and context. They generated semantic descriptions of new gesture categories and provided some preliminary results, but qualitative verification metrics such as recognition accuracy are not clearly given in this paper. Madapana and Wachs\cite{ZSLgesture,madapana2017zsgl} described a new paradigm for Zero-Shot Gestural Learning (ZSGL), in which they generated semantic descriptors for gestures and assessed the performance of various state-of-the-art algorithms. They later proposed another Hard Zero-Shot Learning (HZSL) task of gestures\cite{HZSLgesture} with a small amount of gesture data, and addressed it by integrating One-Shot Learning (OSL) and supervised clustering techniques. However, in the settings of GZSL, the recognition accuracy of their method becomes much lower. More recently, in our previous work\cite{wu2018}, we established a skeletal joint gesture dataset and designed a recognition system for unfamiliar dynamic gestures based on Semantic Auto-Encoder (SAE). This model achieves a much higher accuracy in ZSL setting by utilizing an additional reconstruction constraint, but the issue of the bias towards the seen classes still remains in GZSL setting. The performance of these models in both seen and unseen categories is not satisfactory enough, and thus it is difficult to satisfy the different needs of application and better understand the users' intent.

In order to improve the performance on the GZSL task of hand gesture recognition, we propose an end-to-end prototype-based framework which can mitigate the bias towards predicting seen classes. A detector is first developed to learn gesture representations and discriminate whether the test samples come from the seen or unseen categories. Then, two different classifiers for recognizing seen and unseen classes, respectively, produce the corresponding prediction results for test samples. Inspired by Convolutional Prototype Learning (CPL) framework\cite{CPL} which has shown great potential to handle the open world recognition problem, we use the prototype loss to improve intra-class compactness of the feature representation. In this way, seen samples can be classified to the seen categories which the nearest prototypes belong to, and unseen samples are excluded at the same time via the learned distance threshold. The feature representations of those excluded unseen samples are further taken into the zero-shot label predictor to obtain prediction results. The proposed framework can be trained in an end-to-end manner, which ensures the learning efficiency and makes the feature representations robust to further recognize both seen and unseen categories.

The contributions of this paper are as follows:

\begin{itemize}
\item An end-to-end prototype-based GZSL framework is proposed for hand gesture recognition by integrating prototype learning and feature mapping mechanism. It can improve both the recognition effectiveness and efficiency as all intermediate procedures are updated simultaneously.

\item We build a novel hand gesture dataset which contains 25 hand gestures and 11 semantic attributes for GZSL task. This dataset is extended and re-recorded based on our previous work\cite{wu2018} for hand gesture recognition in ZSL setting.

\item Comprehensive experimental results on our dataset compared with other state-of-the-art methods demonstrate the effectiveness and efficiency of the proposed framework.
\end{itemize}

\section{Related Work}

\subsection{Dynamic Hand Gesture Recognition}
Early works on hand gesture recognition are mainly based on RGB images\cite{signlanguage} or the information captured by data gloves\cite{camastra2013lvq}. However, data gloves tend to be not user-friendly, and RGB images often lose part of the spatial information of hand motion. Recently, with the development of depth sensors, such as the Leap Motion controller (LMC)\cite{lu2016dynamic} and Microsoft Kinect\cite{tang2015real}, rich 3D information of gestures can be obtained, and accurate skeleton data can be more easily extracted. In this paper, we use the LMC to capture skeleton data for its high localization precision.

As for the recognition algorithms, early research on dynamic hand gesture recognition mainly uses hand-crafted features and adopts methods such as Dynamic Time Warping (DTW)\cite{vemulapalli2014human}, Hidden Markov Model (HMM)\cite{signlanguage} and Hidden Conditional Random Field (HCRF)\cite{lu2016dynamic} for classification. With the success of deep learning in a variety of visual tasks, deep networks such as 3D Convolutional Neural Network (3D CNN)\cite{molchanov2016online,hu20183d}, Long Short-Term Memory Networks (LSTM)\cite{zhu2017multimodal,nunez2018convolutional} and graph-based networks\cite{nguyen2019neural} are applied in the field of gesture recognition. Although many approaches have been investigated for hand gesture recognition, the problem of recognizing samples from new unseen classes still remains. A large number of training data are difficult to obtain and the model needs to be retrained after obtaining new data, which limit the practical application of these algorithms in real scenarios. Different from the above works, our proposed method in the more challenging GZSL setting can help address the above problems to better satisfy the different needs of the application.

\subsection{Zero-Shot Learning}
The early works of zero-shot learning directly construct classifiers for seen and unseen class attributes. Lampert et al.\cite{firstZSL} first proposed the task of zero-shot learning and introduced an attribute-based classification approach by leveraging high-level descriptions of object classes to tackle this problem. Later, some other works propose to learn mappings from feature to semantic space. For example, Norouzi et al.\cite{CONSE} mapped images to class embeddings and estimated unseen labels by combining the embedding vectors of the most possible seen classes. Romera-Paredes et al.\cite{ESZSL} developed a simple yet effective approach which learned the relationships between features, attributes and categories by adopting a two-layer linear model. More recently, Kodirov et al.\cite{SAE} adopted an encoder-decoder paradigm that was learned with an additional reconstruction constraint to project a visual feature vector into the semantic space. Morgado and Vasconcelos\cite{SCoRe} proposed two semantic constraints as the complementarity between class and semantic supervision for recognition and achieved state-of-the-art performance. Some other methods further map both features and attributes to a shared space to predict unseen classes. For example, Changpinyo et al.\cite{SYNC} aligned the semantic and feature space by computing the convex combination coefficients of base classifiers to construct classifiers for unseen classes.

\begin{figure*}
\centering
\includegraphics[width=14cm]{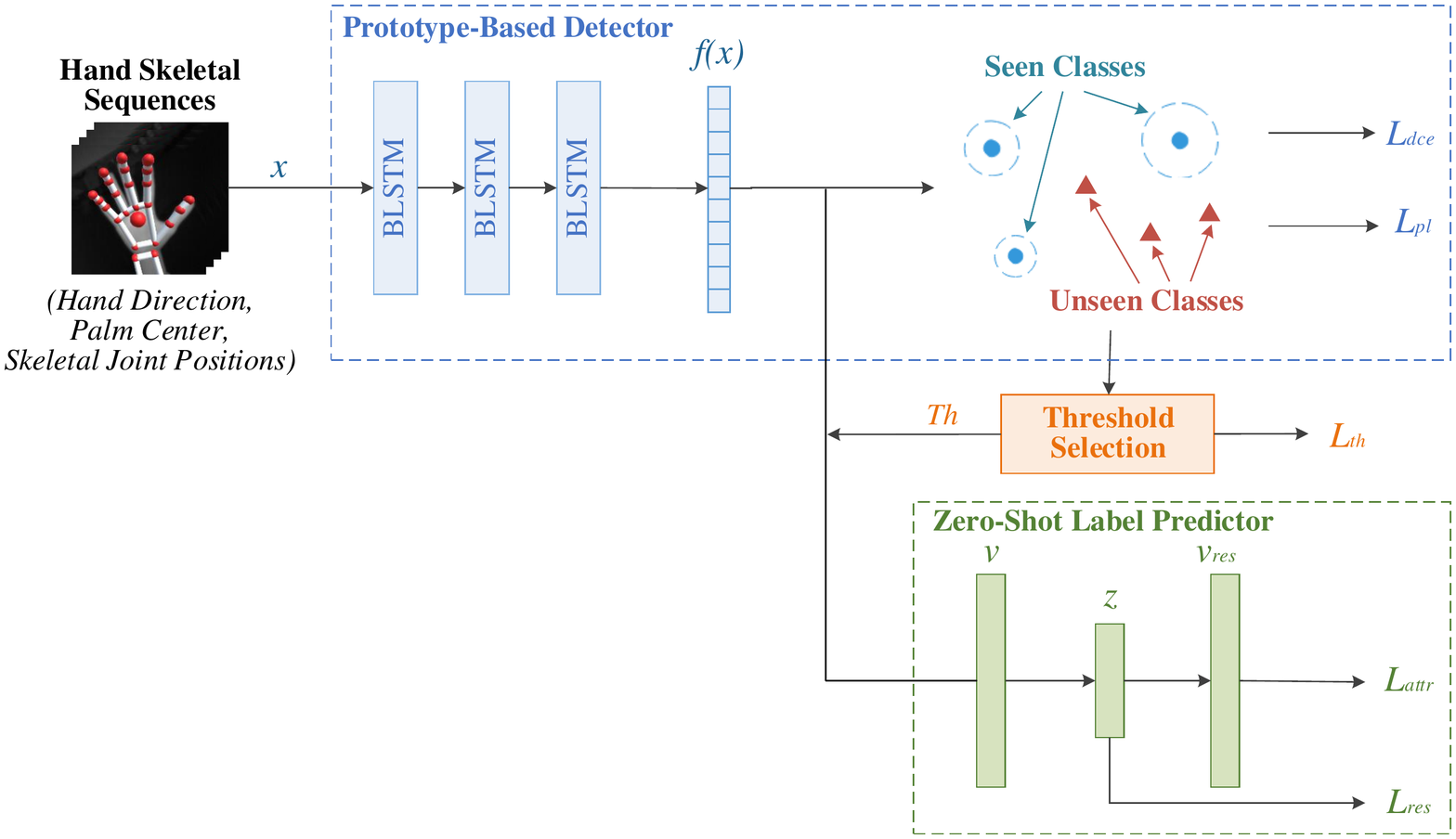}
\caption{Overview of the proposed framework which consists of two branches: a prototype-based detector and a zero-shot label predictor. The first branch takes the gesture sequences as input and outputs the representations in the prototype space. The distance between the representation and the given prototypes determines whether a test sample belongs to a seen or unseen class via the learned threshold. Then, for the samples that are considered to be from unseen categories, their feature representations are further taken as the input to the zero-shot label predictor to obtain recognition results. These two branches can be jointly trained in an end-to-end manner. }
\label{fig1}
\end{figure*}

\subsection{Generalized Zero-Shot Learning}
 The limitation of zero-shot learning is that all test data only come from unseen classes. Therefore, a generalized zero-shot learning setting is proposed where the training and testing classes are not necessarily disjoint by allowing both seen and unseen classes during testing. Recently, many works have been proposed to address this task. For example, Xian et al.\cite{f-CLSWGAN} proposed a generative adversarial network (GAN) that synthesizes CNN features of unseen classes, which are conditioned on class-level semantic information. This generative model alleviates the problem of data imbalance between seen and unseen categories. Another GAN-based model is developed and achieved improvements in balancing accuracy between seen and unseen classes, by combining visual-semantic mapping, semantic-visual mapping and metric learning\cite{GDAN}. Some other approaches formulate this task as a cross-modal embedding problem. For instance, Felix et al.\cite{multi} investigated a multi-modal based algorithm that can balance seen and unseen categories by training both visual and semantic Bayesian classifiers. Schonfeld et al.\cite{CADA-VAE} learned latent features of images and attributes via aligned Variational Autoencoders which contain the essential multi-modal information associated with unseen classes. Although these methods mainly target the challenge of data imbalance between seen and unseen categories, the bias still exists due to the similar treatment of all categories. To solve this, Bhattacharjee et al.\cite{AEdetect} further proposed a novel detector based on an autoencoder with reconstruction and triplet cosine embedding losses to determine whether an input sample belongs to a seen or unseen category. This greatly improves the recognition performance in recognizing novel classes. Mandal et al.\cite{action} further achieved zero-shot action recognition by introducing a separate treatment of seen and unseen action categories and synthesized video features for unseen action categories to train an out-of-distribution detector. The method we propose combines the detector and the seen class classifier as a branch, which has a simpler structure and is more convenient for training.

\section{Methodology}

We first formalize the problem of GZSL, and describe the proposed model for hand gesture recognition, which consists of two modules: a prototype-based detector and a zero-shot label predictor. The prototype-based detection branch first learns a detector that determines whether an input sample belongs to a seen or unseen category, and meanwhile produces feature representations of unseen data. Then, the zero-shot label prediction branch takes these features as input, and outputs predictions of samples from unseen classes through a learned mapping mechanism from feature to semantic space. We then provide the detailed end-to-end learning objective in this section. The proposed framework is shown in Fig.~\ref{fig1}.

\subsection{Problem Definition}

Let $S = \left\{ {{x_s},{y_s},{z_s}\left| {{x_s} \in {\cal X},{y_s} \in {{\cal Y}^s},{z_s} \in {\cal Z}} \right.} \right\}$ be the training data for the seen classes, where ${x_s}$ is the hand skeletal sequence, ${y_s}$ is the label of ${x_s}$ in the set of seen classes ${{\cal Y}^s}$, ${z_s}$ is the corresponding semantic embedding among all embeddings ${\cal Z}$. Similarly, the unseen data can be denoted as $U = \left\{ {{x_u},{y_u},{z_u}\left| {{x_u} \in {\cal X},{y_u} \in {{\cal Y}^u},{z_u} \in {\cal Z}} \right.} \right\}$, where the hand skeletal sequence ${x_u}$ is only available during testing, ${{\cal Y}^u}$ represents the set of unseen labels, and ${{\cal Y}^u} \cap {{\cal Y}^s} = \emptyset $. The goal of GZSL is to learn a classifier $f:{\cal X} \to {{\cal Y}^u} \cup {{\cal Y}^s}$, so that the knowledge can be transferred to recognize samples from both seen and unseen categories.

\subsection{Prototype-Based Detector (PBD)}
\label{sec:PBLSTM}

The prototype-based detector utilizes a multi-layer Bidirectional Long Short-Term Memory Networks (BLSTM)\cite{BLSTM} to extract temporal features from gesture sequences. The input gesture sequences are captured by a Leap Motion Controller, which involve hand direction, palm center and skeletal joint positions. A BLSTM layer is composed of two LSTM layers (a forward one and a backward one), which can capture both past and future contextual information at the same time. Traditionally, a softmax layer is further added on the top of the features extracted by BLSTM for classification. However, this softmax-based approach tends to misclassify unseen to seen classes, thus makes it difficult to distinguish between the seen and unseen categories.

To solve the problem of misclassification, Yang et al.\cite{CPL} proposed a convolutional prototype learning (CPL) framework, which can improve the robustness of classification. CPL aims to learn a few prototypes using CNN features and predict classification labels by matching representations in the prototype space with the closest prototype. Inspired by this, we propose to map the extracted features from BLSTM and learn a fixed number of prototypes for each class. Then, by adding a distance threshold selection process, we can determine whether a test sequence belongs to a trained category.

The BLSTM features of each input $x$ are denoted as $f(x)$, and the parameters of BLSTM are denoted as $\theta $. The features are projected to the prototype space through a FC layer. The projection of $f(x)$ in the prototype space is denoted as $p_{pbd}\left( x \right)$, and the learned prototypes are defined as $M = \left\{ {{m_{ij}}|i = 1, \cdots ,C;j = 1, \cdots ,K} \right\}$, where ${m_{ij}}$ represents the $j^{th}$ prototype of the $i^{th}$ category, $C$ is the number of categories and $K$ is the number of prototypes for each class. The parameters of BLSTM $\theta $ and the prototypes $M$ are jointly trained through the following two loss functions.

The first is the distance-based cross entropy (DCE) loss which is based on the traditional cross entropy loss. It can be defined as:

\begin{equation}
{L_{dce}}\left( {\left( {x,y} \right)\left| {\theta ,M} \right.} \right) =  - \log \sum\limits_{j = 1}^K {\frac{{{e^{ - \gamma dis\left( {p_{pbd}\left( x \right),{m_{yj}}} \right)}}}}{{\sum\nolimits_{k = 1}^C {\sum\nolimits_{l = 1}^K {{e^{ - \gamma dis\left( {p_{pbd}\left( x \right),{m_{kl}}} \right)}}} } }}},
\end{equation}
where $dis\left( {p_{pbd}\left( x \right),{m_{kl}}} \right){\rm{ = }}\left\| {p_{pbd}\left( x \right){\rm{ - }}{m_{kl}}} \right\|_2^2$ computes the distance between $p_{pbd}\left( x \right)$ and ${m_{kl}}$, $\gamma $ is a hyper-parameter. Minimizing the DCE loss helps improve the classification accuracy and enhance the separability among different training classes.

Another prototype loss (PL) is used as a normalization to enhance intra-class compactness, which is defined as:

\begin{equation}
{L_{pl}}\left( {\left( {x,y} \right)\left| {\theta ,M} \right.} \right) = \left\| {p_{pbd}\left( x \right){\rm{ - }}{m_{yj}}} \right\|_2^2,
\end{equation}
where ${m_{yj}}$ is the closest prototype to $p_{pbd}\left( x \right)$ for the class $y$, which can effectively regularize the model and improve the intra-class compactness of the feature representations.

\subsection{Zero-Shot Label Predictor}

In ZSL and GZSL tasks, semantic representations of all seen and unseen categories are available. In order to recognize unseen gestures, a model needs to learn the relationship between the high-level semantic representations and the extracted features $f\left( x \right)$ which are introduced in Section \ref{sec:PBLSTM}. Inspired by the single-layer linear Semantic Auto-Encoder \cite{SAE} for object recognition, we develop a multi-layer Semantic Auto-Encoder (SAE) as the classifier to improve the prediction results by stacking more layers.

We use fully connected (FC) layers with symmetric structure as the encoder and decoder of the SAE. All the parameters of the encoder and decoder are denoted as $\phi $. The input and the output of the encoder are denoted as $v{\rm{ = }}f\left( x \right)$ and $z$, and the output of the decoder is denoted as ${v_{res}}$. The SAE aims to learn a mapping, which projects the learned representations from feature space to semantic space. The mapped semantic embedding $z$ is trained to be close to the given semantic prototype of the corresponding category, and at the same time, the SAE can retain the original input information through the reconstruction of the decoder. The loss function of SAE consists of an attribute loss ${L_{attr}}$ and a reconstruction loss ${L_{res}}$ as:

\begin{equation}
{L_{attr}}\left( {\left( {x,{z_s}} \right)\left| {\theta ,\phi } \right.} \right) = \left\| {z - {z_s}} \right\|_2^2,
\end{equation}
\begin{equation}
{L_{res}}\left( {\left( {x,{z_s}} \right)\left| {\theta ,\phi } \right.} \right) = \left\| {v - {v_{res}}} \right\|_2^2,
\end{equation}
where $\theta$ represents the parameters for BLSTM in our model.

\subsection{End-to-End Learning Objective}

The two branches which integrate feature extraction and label prediction can be jointly trained in an end-to-end manner. Therefore, both the parameters of prototype-based detector and SAE are learned at the same time. Thus, the joint learning objective of our end-to-end framework can be formulated as:

\begin{equation}
L\left( {\left( {x,y,{z_s}} \right)\left| {\theta ,M,\phi } \right.} \right) = {L_{dce}} + {\lambda _1}{L_{pl}} + {\lambda _2}{L_{attr}} + {\lambda _3}{L_{res}},
\end{equation}
where ${\lambda _1}$, ${\lambda _2}$, ${\lambda _3}$ are hyper-parameters which weight the above four loss terms.

After training the end-to-end network, the distance thresholds for visual prototypes are further set, in order to determine whether a new sample belongs to a seen or unseen categories. Specifically, we fix the trained network parameters and prototypes, and use all the training data $X_{tr}=\left\{ x_i|i = 1, \cdots ,N_{tr} \right\}$ to learn the thresholds $Th = \left\{ {{th_{kl}}|k = 1, \cdots ,C;l = 1, \cdots ,K} \right\}$, where ${x_i}$ is the $i^{th}$ gesture sequence in the training set, $N_{tr}$ is the number of training samples, and $th_{kl}$ is the corresponding threshold of $m_{kl}$. The loss function is given by:

\begin{equation}
{L_{th1}}\left( {X_{tr}|Th} \right) = \frac{1}{N_{tr}}\sum\limits_{i = 1}^{N_{tr}} {\left\{ {\begin{array}{*{20}{c}}
{0,\Delta d\left( x_i \right) \le  0}\\
{\Delta d\left( x_i \right) + 1,\Delta d\left( x_i \right) > 0},
\end{array}} \right.}
\end{equation}
\begin{equation}
{L_{th2}}\left( {X_{tr}|Th} \right) = \left\| {Th} \right\|_2^2,
\end{equation}
\begin{equation}
{L_{th}}\left( {X_{tr}|Th} \right) = {L_{th1}}\left( {X_{tr}|Th} \right) + \beta {L_{th2}}\left( {X_{tr}|Th} \right),
\end{equation}
where $\Delta d\left( x_i \right) = {d_m\left( x_i \right)} - {Th\left( x_i \right)}$, ${d_m\left( x_i \right)}$ represents the minimal distance between $p_{pbd}\left( {x_i} \right)$ and all prototypes, and $Th\left( x_i \right)$ is the corresponding threshold of the closest prototype of $x_i$. ${L_{th1}}$ aims to correctly classify the samples that belong to the current category, and ${L_{th2}}$ is used as the regularization to reduce the influence of outliers on thresholds. By using the hyper-parameter $\beta $ that weights the above two parts, the model is able to learn the thresholds that can best discriminate samples of different categories.

\subsection{Label Prediction}

During predicting, an test sample $x$ will be classified into the category of the closest prototype by the prototype-based detector:

\begin{equation}
\varepsilon \left( x \right) = \mathop {\arg \min }\limits_{i = 1}^C \left( {\mathop {\min }\limits_{j = 1}^K \left\| {p_{pbd}\left( x \right){\rm{ - }}{m_{ij}}} \right\|_2^2} \right),
\end{equation}
where $\varepsilon \left( x \right)$ represents the category which the closest prototype belongs to. Then, the model distinguishes the seen and unseen categories by comparing the minimum distance with the thresholds:

\begin{equation}
{d_m}\left( x \right) = \mathop {{\rm{min}}}\limits_{i = 1}^C \left( {\mathop {\min }\limits_{j = 1}^K \left\| {p_{pbd}\left( x \right){\rm{ - }}{m_{ij}}} \right\|_2^2} \right),
\end{equation}
\begin{equation}
label_{pbd}\left( x \right) = \left\{ {\begin{array}{*{20}{c}}
{\varepsilon \left( x \right),{d_m}\left( x \right)  \le  Th\left(x \right)}\\
{{\rm{unseen}},{d_m}\left( x \right) > Th\left( x \right)},
\end{array}} \right.
\end{equation}
where $label_{pbd}\left( x \right)$ represents the intermediate prediction results of the prototype-based detector, and $Th\left( x \right)$ is the threshold of the closest prototype of $x$.

Then, for the sample which is considered to be from an unseen category, its feature $f\left( x \right)$ is further projected into the semantic space by the SAE. We compare the projected semantic representation $z$ with the semantic prototypes of all unseen classes. The zero-shot prediction result $\varepsilon_u\left( x \right)$ is given by:

\begin{equation}
\varepsilon_u\left( x \right) = \mathop {\arg \min }\limits_j \left\| {z{\rm{ - }}Z_j^u} \right\|_2^2,
\end{equation}
where $Z_j^u$ is the semantic prototype of the $j^{th}$ unseen category.

In summary, the prediction result of a test sample is as follows:

\begin{equation}
label\left( x \right) = \left\{ {\begin{array}{*{20}{c}}
{\varepsilon \left( x \right),{d_m}\left( x \right) \le  Th\left( x \right)}\\
{\varepsilon_u\left( x \right),{d_m}\left( x \right) > Th\left( x \right)}.
\end{array}} \right.
\end{equation}

\section{Experiments}

\subsection{Experiment Settings}

\subsubsection{Dataset}

\begin{figure}
\centering
\includegraphics[width=8.5cm]{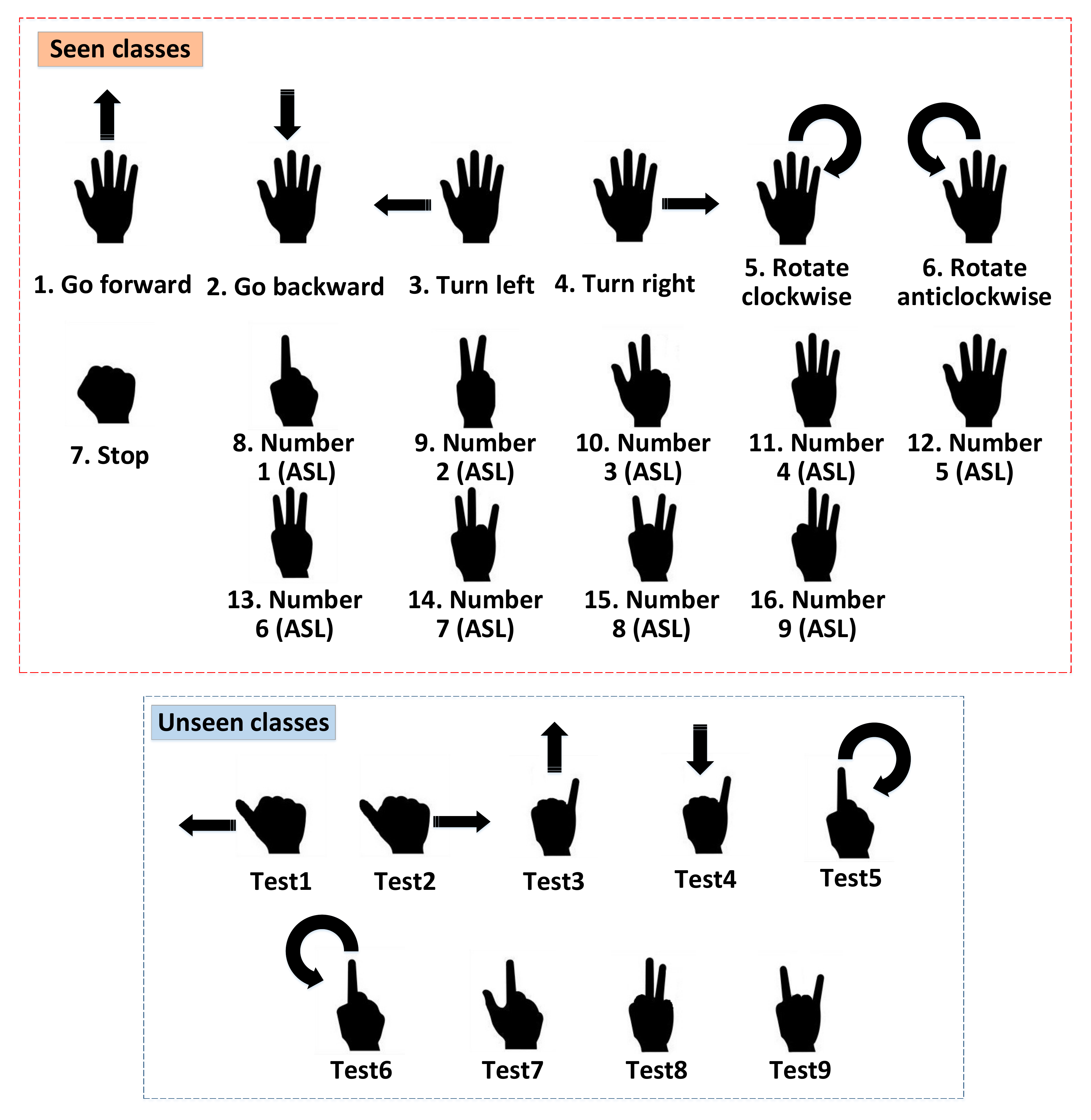}
\caption{Hand gestures in our dataset.}
\label{fig2}
\end{figure}

\begin{figure}
\centering
\includegraphics[width=8cm]{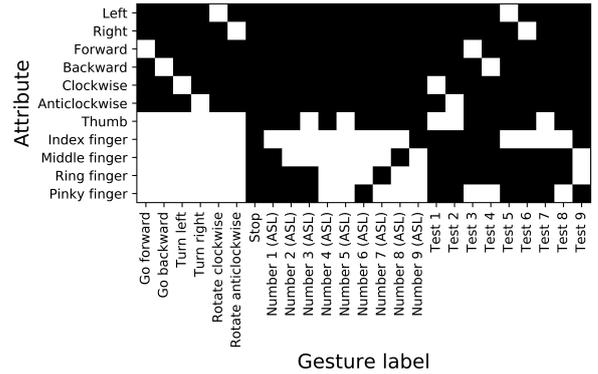}
\caption{Binary heat map of the categories and attributes.}
\label{fig3}
\end{figure}

This dataset is an expansion of the dataset proposed in our previous work\cite{wu2018} which contains 16 seen gestures in the training set and 4 unseen gestures in the test set. As shown in Fig.~\ref{fig2}, 16 seen gestures and 9 unseen gestures are developed in our novel dataset, which are captured by a Leap Motion Controller. More differences in hand position and gesture custom are taken into account. In total, 800 sequences of all seen categories are contained in the training set, and 500 sequences of both seen and unseen categories are contained in the test set. Each sequence consists of 100 frames. Meanwhile, the information such as hand direction, palm center and skeletal joint positions on a single right hand is recorded for each frame. To better extract hand posture and relative motion information, the recorded data is preprocessed and normalized. We further design 11 attributes including hand movement and finger bending states for each category based on the experience of gesture recognition research. All attributes are binary and they are visualized in the form of heat map which are shown in Fig.~\ref{fig3}.

\begin{table*}
\caption{The Experimental Results of the State-of-the-art Comparisons in GZSL Setting}
\centering
\tabcolsep5pt
\renewcommand\arraystretch{1.6}
\begin{tabular}{l|c|c|c}
\hline
Methods              & $Ac{c_s}$   & $Ac{c_u}$    & $H$       \\ \hline
ESZSL \cite{madapana2017zsgl}            & 77.81\% & 13.89\% & 23.57\% \\
CADA-VAE \cite{CADA-VAE}            & 80.00\% & 53.89\% & 64.40\% \\
f-CLSWGAN \cite{f-CLSWGAN}            & 79.79\% & 55.00\% & 65.08\% \\
End-to-End Framework (Ours)  & \textbf{89.06\%} & \textbf{58.33\%} & \textbf{70.49\%} \\ \hline
\end{tabular}
\label{table1}
\end{table*}

\begin{table*}
\caption{The Experimental Results of the Ablation Analysis}
\centering
\tabcolsep5pt
\renewcommand\arraystretch{1.6}
\begin{tabular}{l|c|c|c|c}
\hline
Methods              & $Ac{c_s}$   & $Ac{c_u}$   & $H$    & Test Time  \\ \hline
BLSTM+SAE\cite{SAE}            & \textbf{91.88\%} & 15.00\% & 25.79\% & 0.023s\\
End-to-End Framework (Fixed Threshold)    & 84.69\% & 50.56\% & 63.31\% & 0.022s \\
PBD+SAE             & 90.63\% & 57.22\% & 70.15\% & 0.026s \\
End-to-End Framework  & 89.06\% & \textbf{58.33\%} & \textbf{70.49\%} & \textbf{0.022s} \\ \hline
\end{tabular}
\label{table2}
\end{table*}

\subsubsection{Evaluation Metrics}

We adopt the top-1 accuracy to evaluate the models, and the top-1 accuracy of seen classes and unseen classes are denoted as $Ac{c_s}$ and $Ac{c_u}$. As there is an inherent bias towards the seen classes, and to ensure that both $Ac{c_s}$ and $Ac{c_u}$ are high enough, we use harmonic mean $H$ for the final performance comparison, which can be defined as:

\begin{equation}
H{\rm{ = }}\frac{{2 \times Ac{c_s} \times Ac{c_u}}}{{Ac{c_s}{\rm{ + }}Ac{c_u}}}.
\end{equation}

\subsubsection{Implementation Details}

We utilize a three-layer BLSTM network to extract features, and the numbers of forward and backward LSTM neurons are set to 64. The features are mapped to prototype space through a fully connected layer. We maintain one prototype for each category, and the dimension of prototypes is set to 20. The activation function we use in the network is ReLU. For SAE, both encoder and decoder have two hidden layers. The input dimension of the encoder is the same as the feature dimension, which is 128. The output dimension of the encoder is the same as the number of attributes, which is 11. During training, the batch size is set to 8, the learning rate is set to 0.001, and the number of training epochs is set to 100. The Adam optimizer\cite{adam} is utilized to minimize the loss. Other hyper-parameters are selected by 10-fold cross-validation. $\beta $ of threshold selection is set to 0.1, and ${\lambda _1},{\lambda _2},{\lambda _3}$ of the end-to-end learning objective are set to 5, 5 and 0.05, respectively.

\subsection{State-of-the-art Comparisons}

In this section, we compare our proposed framework in the GZSL setting with one of the state-of-the-art methods for zero-shot gesture recognition proposed in \cite{madapana2017zsgl}, which utilized three ZSL methods and obtained the best recognition results using Embarrassingly Simple Zero-Shot Learning (ESZSL)\cite{ESZSL}. In order to better demonstrate the effectiveness of our algorithm, we also choose two state-of-the-art methods for object recognition, CADA-VAE\cite{CADA-VAE} and f-CLSWGAN\cite{f-CLSWGAN}, for the comparisons on our dataset at the same time. The features of the seen and unseen categories are obtained by a three-layer BLSTM network. Experimental results are shown in Table \ref{table1}.

We observe that our proposed method outperforms the state-of-the-art methods, and $Ac{c_s}$, $Ac{c_u}$ and $H$ are increased by 9.06\%, 3.33\% and 5.41\%, respectively. The recognition accuracy is decreased for CADA-VAE and f-CLSWGAN because the two methods predict labels of both seen and unseen categories by using only a single model. Specifically, it is difficult to maintain the recognition accuracy of the seen category while considering the generalization performance. In our work, however, two separate classifiers are used for label prediction, thus the generalization ability is enhanced to reduce the impact of bias. In addition, the complexity of their models is higher than ours, and more parameters that need to be learned require more training data which can be inefficient and unavailable.

\subsection{Ablation Analysis}

We analyze different components in our framework including the prototype-based detector, threshold selection and end-to-end training manner.

\textbf{Prototype-Based Detector.} We choose the traditional SAE \cite{SAE} without the prototype-based detector as a baseline of the GZSL task of hand gesture recognition. The features are first obtained by a three-layer BLSTM network. We then feed them into the traditional SAE for predictions of both seen and unseen categories. The experimental result is shown in Table \ref{table2} (line 1). We observe that although the traditional SAE performs slightly better on $Ac{c_s}$, there is a severe bias towards the seen categories. Our framework combining prototype-based detector and SAE achieves the improvement of 44.7\% on harmonic mean over the traditional SAE model. This demonstrates that the prototype-based detector can effectively separate the unseen category from the seen category, which greatly reduces the impact of learning bias.

\textbf{Threshold Selection.} In order to verify the effectiveness of our threshold selection method, we compare it to the method with a fixed threshold for all seen categories and explore the impact of the parameter $\beta $ on the discrimination between the seen and unseen categories. Except for the threshold selection part, other modules of the different comparison models are identical. The Comparison Results of Acceptance Rate (AR) and Rejection Rate (RR) for Different Threshold Selection are shown in Table \ref{table3}. AR denotes the percentage of the accepted samples in the test samples from the seen categories, while RR denotes the percentage of the rejected samples in the test samples from the unseen categories. The results demonstrate that our method can achieve effective trade-off regarding the acceptance rate and rejection rate, and thus enhance the ability of modeling the cross-class difference as well as the intra-class consistency. Based on the best parameter selection where the fixed threshold is set to 0.5 and $\beta $ of threshold selection is set to 0.01, the recognition results of the seen and unseen categories can be seen in Table \ref{table2} (line 2 and line 4, respectively). We observe that our threshold selection method has a better performance in $Ac{c_s}$, $Ac{c_u}$ and $H$, because improper selection of thresholds can lead to misclassification of both the seen and unseen categories.

\begin{table}
\caption{The Comparison Results of AR and RR for Different Threshold Selection}
\centering
\tabcolsep5pt
\renewcommand\arraystretch{1.6}
\begin{tabular}{c|c|c|c|c|c}
\hline
\multicolumn{3}{c|}{Fixed Threshold} & \multicolumn{3}{c}{Our Method} \\ \hline
Th        & AR          & RR          & $\beta $    & AR        & RR        \\ \hline
0.01      & 64.69\%     & 97.22\%     & 0.5     & 76.56\%   & 96.11\%   \\
0.05      & 81.56\%     & 92.22\%     & 0.2     & 83.12\%   & 93.33\%   \\
0.1       & 85.00\%     & 83.33\%     & 0.05    & 87.50\%   & 86.11\%   \\
0.2       & 87.19\%     & 81.11\%     & 0.02    & 91.00\%   & 77.22\%   \\
0.5       & 90.81\%     & 63.33\%     & 0.01    & 93.12\%   & 72.00\%   \\
1         & 95.93\%     & 48.88\%     & 0.005   & 95.62\%   & 66.11\%   \\ \hline
\end{tabular}
\label{table3}
\end{table}

\textbf{End-to-End Training Manner.} We also compare the performance of our end-to-end framework to the framework where two branches are trained separately. In the third line of Table \ref{table2}, we can observe that when the framework is trained separately, the performance for seen classes is slightly increased compared to our end-to-end framework. This is because although jointly training makes the feature extraction more suitable for both prototype-based detector and SAE, it is difficult to fully satisfy these two tasks using a small amount of data. However, the end-to-end network has the advantage of higher speed and better performance for unseen classes. It is more suitable for the tasks that require real-time performance while ensuring recognition accuracy.

\section{Conclusion}

In this paper, we propose a prototype-based GZSL framework for hand gesture recognition. Two branches of our framework are introduced: a prototype-based detector and a zero-shot label predictor. The prototype-based detector combines feature extraction and prototype learning, which can determine whether a test sample belongs to an unseen category and obtain the prediction results of the samples from the seen categories. In the zero-shot label prediction branch, the SAE is utilized to learn the mapping from the feature space to the semantic space and further predict labels for the samples from the unseen categories. In addition, we design a joint learning objective to train the entire framework in an end-to-end manner. We establish a dataset for evaluating this GZSL task of hand gesture recognition, and the experimental results demonstrate that the proposed framework achieves a significant improvement over the state-of-the-art methods. In future work, we aim to extend this framework to a larger scale of gesture data in order to better support human-robot interaction in the real world.

\section*{Acknowledgment}

This work is supported by the National Natural Science Foundation of China (Grant No. 61673378) and the Ministry of Science and Technology of the People's Republic of China (Grant No. 2017YFC0820200).



\end{document}